\documentclass[journal]{IEEEtran}
\IEEEoverridecommandlockouts

\usepackage{amsmath,amssymb,amsthm,amsfonts}
\usepackage{bm}

\usepackage{algorithm}
\usepackage{algpseudocode}
\usepackage{graphicx}
\usepackage{subfig}
\usepackage{booktabs}
\usepackage{cite}
\usepackage{url}
\usepackage{xcolor}
\usepackage{hyperref}
\usepackage{tcolorbox}

\newtheorem{theorem}{Theorem}
\newtheorem{lemma}{Lemma}

\newcommand{\C}{\mathbb C}

\hypersetup{
  colorlinks=true,
  linkcolor=blue,
  citecolor=blue,
  urlcolor=blue,
}

\newcommand{\vct}[1]{\bm{#1}}     
\newcommand{\mtx}[1]{\bm{#1}}     

\title{The Optimization Landscape of Carath\'eodory Decomposition of Toeplitz Covariances}

\author{Daniel~Busbib~and~Ami~Wiesel,~\IEEEmembership{Senior~Member,~IEEE}
\thanks{D. Busbib and A. Wiesel are with the School of Computer Science and Engineering, The Hebrew University of Jerusalem, Jerusalem 91904, Israel (e-mail: daniel.busbib@mail.huji.ac.il; ami.wiesel@mail.huji.ac.il). A preliminary version of this project was presented at IEEE CAMSAP \cite{busbib2025toeplitz}.}
\thanks{This work was supported in part by the Israel Science Foundation under Grant 2963/25.}}

\begin{document}
\maketitle

\begin{abstract}
Toeplitz covariance estimation is a classical signal processing problem, yet the geometry of the Gaussian maximum-likelihood objective remains only partially understood. Recent Newton-type, majorization-minimization, and gradient-based methods often solve the nonconvex problem globally with enough samples, but also reveal a difficult computational landscape. We study this phenomenon through an overparameterized Carath\'eodory representation, which parameterizes the covariance by the frequencies and amplitudes of steering vectors. A simple example shows that, for any uniform frequency grid, optimizing only the amplitudes can leave a positive error even in the population setting, which motivates optimizing the frequencies as well. Our main result proves that this nonlinear parameterization adds no stationary points: for any sample covariance, every stationary point with a positive definite covariance gives a stationary point of the Toeplitz maximum-likelihood problem. In the population setting, every such point therefore recovers the true covariance. We also prove that gradient descent converges to the true covariance in the population setting, and to the unique maximum likelihood estimate when the sample covariance is close enough to the true covariance. These results do not require overparameterization. Numerically, overparameterization speeds up gradient descent and gives a more accurate estimate for a fixed number of iterations.
\end{abstract}


\section{Introduction}

Covariance estimation is a core task in statistical signal processing with applications across radar detection, hyperspectral imaging, and modern learning systems \cite{kay1998fundamentals, wiesel2015structured, 8052571, 91176, 6678280, li1999computationally, mestre2008asymptotic, smith2005covariance}. When observations come from a wide-sense stationary process, the covariance between two samples depends only on their lag, which implies that the covariance matrix has a Toeplitz structure \cite{kay1998fundamentals, stoica2005spectral}.

There is a rich literature on Toeplitz covariance estimation. Low-complexity estimators include diagonal averaging \cite{eldar2019sampleefficientToeplitzcovariance}, FFT-based circulant approximations \cite{91176, 1458092, zhu2017asymptoticequivalencecirculanttoeplitz}, and tapering \cite{Cai_2010, cai2013toeplitz}. Geometric methods compute the Toeplitz matrix closest to the sample covariance, for example with Dykstra's projections \cite{9054969}. Maximum likelihood methods include expectation-maximization \cite{1458092, turmon1994maximum}, majorization-minimization \cite{7565553, ATOM24}, a Newton-type method that uses fast Fourier transforms to compute the gradient and Hessian \cite{cederberg2024toeplitz}, and a method based on the Gohberg--Semencul factorization \cite{boeck2025gohbergsemencul}. Similar nonconvex techniques were used for other covariance structures, such as sparse and factor models \cite{wei2023large, zou2024factor}. The maximum likelihood methods above perform well when there are enough samples, but they rely on sophisticated, problem-specific optimization machinery. We instead analyze the optimization landscape of a possibly overparameterized Carath\'eodory decomposition and the convergence of gradient descent (GD).

Maximum likelihood estimation in linear covariance models, a class that includes Toeplitz covariances, is nonconvex and may have many local maxima, but local ascent started at a suitable point finds the global maximum with high probability when the number of samples is large enough \cite{zwiernik2017maximum}. These results are for the linear parameters of the model, which for Toeplitz covariances are the lags. The Carath\'eodory decomposition describes the same matrices, but its frequencies enter nonlinearly, and such a change of parameters can add stationary points. We show that this does not happen. For the Toeplitz case, recent numerical studies report that with enough samples, whenever their optimization over the covariance lags ends at a positive definite matrix, it reaches the same solution from different initializations \cite{abramovich2026numerical, abramovich2026properties}. Our results explain this behavior when the sample covariance is close enough to the true covariance.

Toeplitz covariance estimation is closely related to spectral estimation and direction-of-arrival (DOA) problems. The link is via the Carath\'eodory decomposition, which states that any $P\times P$ positive definite Toeplitz covariance matrix can be represented as a non-negative combination of rank-one outer products of steering vectors plus a noise floor \cite{caratheodory1911harmonic, grenander1958toeplitz, stoica2005spectral}:
\begin{equation}
\mtx{C}_{P\times P} = \sum_{k=1}^{K} a_k \vct{v}(\omega_k) \vct{v}(\omega_k)^{\mathrm{H}} + \sigma^2 \mtx{I}_P,
\label{eq:caratheodory_intro}
\end{equation}
where $\vct{a}\in\mathbb{R}_+^K$ and $\vct{\omega}\in\mathbb{R}^K$ contain the amplitude and frequency parameters, respectively, $\sigma^2\geq0$ is a noise variance, and 
\begin{equation}
\vct{v}(\omega) = \begin{bmatrix} 1 & e^{i\omega} & \cdots & e^{i\omega(P-1)} \end{bmatrix}^{\top}
\end{equation}
is the steering vector at frequency $\omega$. This is the classical Carath\'eodory--Fej\'er (Vandermonde) decomposition: for a positive definite Toeplitz $\mtx{C}$, choosing the noise level $\sigma^2$ as the smallest eigenvalue of $\mtx{C}$ makes $\mtx{C}-\sigma^2\mtx{I}$ a rank-deficient positive semidefinite Toeplitz matrix of rank $K\leq P-1$, which has a \emph{unique} representation $\sum_{k=1}^{K} a_k\,\vct{v}(\omega_k)\vct{v}(\omega_k)^H$ with distinct frequencies $\omega_k\in[0,2\pi)$ and strictly positive amplitudes $a_k$. The noise term can also be absorbed into the atoms, as we do in Section~\ref{sec:param}. When the signal consists of a few uncorrelated sources in white noise, the decomposition provides physical interpretability: each rank-one term corresponds to a distinct source~\cite{yang2015gridless}.

Seminal works in spectral estimation include Capon's method \cite{capon1969high} and subspace algorithms such as MUSIC \cite{schmidt1986multiple} and ESPRIT \cite{roy1989esprit}, which estimate the source frequencies from the covariance. Closer to this paper are SPICE \cite{stoica2011spice} and its weighted variant WSPICE \cite{Stoica_2014_wspice} that bridge covariance fitting and spectral sparsity. These methods rely on overparameterized models with $K\gg P$ frequencies. However, they assume a fixed grid of frequencies $\omega_k$ and only optimize $a_k$. A related line of work avoids the grid entirely. Gridless methods based on atomic norm minimization and super-resolution recover off-grid frequencies by solving convex semidefinite programs \cite{tang2013compressed, candes2014superresolution, yang2015gridless}. Other works estimate the covariance by maximum likelihood and then extract the frequencies \cite{pote2023maximum}, or find the frequencies directly by gradient descent \cite{fannjiang2026gradient}. In standard covariance estimation tasks, the focus is only on recovering the overall covariance accurately rather than identifying the specific physical frequencies. In this context, overparameterized models sacrifice identifiability and interpretability in the hope of achieving better optimization.

The main motivation for this paper is the recent progress on the global optimality of overparameterized GD in nonconvex optimization. This phenomenon is often viewed as one reason for the success of modern deep learning \cite{du2018gradient,liu2022loss,oymak2019overparameterized,xu2025localpolyaklojasiewiczdescentlemma}. Similar results were also proved beyond neural networks, for example in low-rank matrix factorization \cite{zhang2024improved} and in sparse optimization over measures \cite{chizat2022sparse}. Briefly, it has been shown that when the model is sufficiently overparameterized, GD often converges to a global minimizer close to its initialization. These results suggest that overparameterization can improve the behavior of gradient descent and naturally raise the question of whether overparameterization can also help in nonconvex covariance estimation.

The main contributions are summarized as follows:

\begin{itemize}
\item \textbf{No new stationary points.}
We prove that for every $K\geq P$, if a stationary point of the joint amplitude-frequency objective gives a positive definite covariance, then this covariance is a stationary point of the Toeplitz likelihood. In the population setting, every such point recovers the true covariance. When the sample covariance is close enough to the true covariance, the Toeplitz likelihood has a unique stationary point, and the covariance iterates of gradient descent converge to it.

\item \textbf{Joint amplitude-frequency gradient descent.}
We propose a simple gradient descent framework for Toeplitz covariance estimation. The covariance is parameterized by an overparameterized Carath\'eodory decomposition, and both the amplitudes and the frequencies are optimized by minimizing the Gaussian negative log-likelihood.

\item \textbf{Benefits of overparameterization.}
We show empirically that overparameterization speeds up the convergence of gradient descent, so that for a fixed number of iterations it yields a more accurate estimate, while the final MSE is similar for all tested values of $K$ and close to the Cram\'er--Rao bound.
\end{itemize}

\subsection*{Notation}

We use normal letters ($P, K, a, \omega$) to denote scalars, bold lowercase letters ($\vct{x}, \vct{a}, \vct{\omega}$) for vectors, and bold uppercase letters ($\mtx{S}, \mtx{C}$) for matrices. We use $\|\cdot\|$ for the Euclidean norm of a vector, $\|\cdot\|_F$ and $\langle\cdot,\cdot\rangle_F$ for the Frobenius norm and inner product, $\|\cdot\|_2$ for the spectral norm, and $\operatorname{tr}(\cdot)$ and $\det(\cdot)$ or $|\cdot|$ for trace and determinant. The superscripts $(\cdot)^T$ and $(\cdot)^H$ denote the transpose and the conjugate transpose, and for Hermitian matrices $\mtx{A}\succ\mtx{B}$ and $\mtx{A}\succeq\mtx{B}$ mean that $\mtx{A}-\mtx{B}$ is positive definite and positive semidefinite, respectively. We write $\mathbb{E}[\cdot]$ for expectation, and vector inequalities such as $\vct{a}\geq \vct{0}$ are componentwise.

\section{Problem Formulation}\label{sec:problem}

\subsection{Statistical Model and Maximum Likelihood}

We consider the problem of estimating an unknown Toeplitz covariance matrix from independent realizations of a zero-mean stationary proper complex Gaussian signal. Let $\vct{x} \in \mathbb{C}^P$ satisfy
\begin{align}
    &\mathbb{E}[\vct{x}]=\vct{0}\nonumber\\
    &\mathbb{E}[\vct{x}\vct{x}^{\top}]=\mtx{0}\nonumber\\
    &\mathbb{E}[\vct{x}\vct{x}^{\mathrm H}]=\mtx{C}\in {\mathcal{T}},
\end{align}
where ${\mathcal{T}}$ is the set of
 $P\times P$ Hermitian positive definite Toeplitz covariance matrices:
\begin{align}\label{Toeplitz_set}
 \mathcal{T} = \{ \mtx{C}\succ 0:\; \mtx{C}=\mtx{C}^{\mathrm H},\; \mtx{C}_{ij}=c_{i-j}\ \text{for some scalars } c_{\ell}\} 
\end{align}
Apart from the Toeplitz structure, no prior information on $\mtx{C}$ is assumed.

We observe $M$ independent and identically distributed realizations of $\vct{x}$, denoted by $\vct{x}_m$ for $m=1,\ldots,M$. Under the Gaussian model, the sample covariance $\mtx{S}$ is a sufficient statistic:
\begin{equation}
    \mtx{S} = \frac{1}{M} \sum_{m=1}^M \vct{x}_m \vct{x}_m^{\mathrm H}.
\end{equation}  
The maximum likelihood estimate (MLE) minimizes the per-sample Gaussian negative log-likelihood (NLL), up to an additive constant,
\begin{equation}
\mathrm{NLL}(\mtx{C}) = \operatorname{tr}(\mtx{S}\,\mtx{C}^{-1}) + \log|\mtx{C}|,
\label{eq:nll_C}
\end{equation}
over $\mtx{C}\in\mathcal{T}$. For a singular $\mtx{S}$ this minimum need not be attained \cite{fuhrmann1988existence}.

\subsection{Carath\'eodory Parameterization}\label{sec:param}

Rather than optimizing over $\mathcal{T}$ directly, we parameterize the candidate covariance as a non-negative combination of steering-vector outer products,
\begin{equation}
\hat{\mtx{C}}(\vct{a},\vct{\omega}) = \sum_{k=1}^{K} a_k\, \vct{v}(\omega_k)\vct{v}(\omega_k)^{\mathrm{H}},
\qquad a_k\geq 0,\ \omega_k\in[0,2\pi),
\label{eq:param_model}
\end{equation}
where $\vct{v}(\omega)$ is the steering vector. The number of atoms $K$ controls the size of the parameterization. A positive definite $\hat{\mtx{C}}$ needs at least $P$ atoms, so $K=P$ is the smallest useful choice, and $K>P$ corresponds to an overparameterized regime. By construction, $\hat{\mtx{C}}(\vct{a},\vct{\omega})$ is Hermitian Toeplitz and positive semidefinite.

Unlike \eqref{eq:caratheodory_intro}, the model \eqref{eq:param_model} has no separate noise term $\sigma^2\mtx{I}$. This is not a limitation when $K\geq P$, because the noise can itself be written with atoms. For example, the identity is the average of $P$ atoms on a uniform grid, $\mtx{I}_P=\frac{1}{P}\sum_{k=0}^{P-1}\vct{v}(2\pi k/P)\vct{v}(2\pi k/P)^{\mathrm{H}}$. More generally, every $\mtx{C}\in\mathcal{T}$ can be written with at most $P$ atoms and no noise term. To see this, pick any frequency $\omega$ and subtract from $\mtx{C}$ the largest multiple of the atom $\vct{v}(\omega)\vct{v}(\omega)^{\mathrm{H}}$ that keeps the result positive semidefinite. This multiple is $\mu=1/(\vct{v}(\omega)^{\mathrm{H}}\mtx{C}^{-1}\vct{v}(\omega))$, and what remains is a singular positive semidefinite Toeplitz matrix. By the Carath\'eodory--Fej\'er decomposition, such a matrix is a combination of at most $P-1$ atoms, so $\mtx{C}$ is a combination of at most $P$ atoms \cite{georgiou2007caratheodory, stoica2005spectral}.

Since $\omega$ was arbitrary, this gives infinitely many decompositions of the same $\mtx{C}$, and with $K>P$ there are even more, since an atom can be split into several atoms at the same frequency. Different parameters can therefore give the same covariance, which is why our results are stated for the covariance $\hat{\mtx{C}}$ and not for the parameters.

Substituting \eqref{eq:param_model} into the NLL \eqref{eq:nll_C} gives the objective
\begin{equation}
L(\vct{a},\vct{\omega}) = \operatorname{tr}\!\left(\mtx{S}\,\hat{\mtx{C}}(\vct{a},\vct{\omega})^{-1}\right) + \log\bigl|\hat{\mtx{C}}(\vct{a},\vct{\omega})\bigr|,
\label{eq:objective}
\end{equation}
which we minimize over $(\vct{a},\vct{\omega})$.

\subsection{Limitations of a Fixed Frequency Grid}\label{sec:grid}

A natural simplification of \eqref{eq:param_model} is to fix the frequencies on a grid and optimize only the amplitudes, as in SPICE-type methods \cite{stoica2011spice,Stoica_2014_wspice}. It is well known that a fixed grid cannot represent every Toeplitz covariance, which is the reason gridless methods were developed \cite{ATOM24, yang2015gridless}.

Failure occurs even for a single off-grid source. Fix $P\geq2$ and $K\geq2$, take the uniform grid $\omega_k=2\pi(k-1)/K$, $k=1,\ldots,K$, and consider a single source halfway between two grid points, $\mtx{C}=b\,\vct{v}(\omega_0)\vct{v}(\omega_0)^{\mathrm{H}}+\sigma^2\mtx{I}$ with $\omega_0=\pi/K$, $b>0$ and $\sigma^2>0$. For any amplitudes $\vct{a}\geq\vct{0}$, the matrix $\hat{\mtx{C}}$ in \eqref{eq:param_model} with these frequencies has the entries $\hat r_0=\sum_k a_k$ on its main diagonal and $\hat r_1=\sum_k a_k e^{i\omega_k}$ on its first subdiagonal. Since $\omega_k-\omega_0=(2k-3)\pi/K$ is an odd multiple of $\pi/K$, we have $\cos(\omega_k-\omega_0)\leq\cos(\pi/K)$ for every $k$, and therefore
\begin{equation}
\operatorname{Re}\!\left(e^{-i\omega_0}\hat r_1\right)=\sum_{k=1}^K a_k\cos(\omega_k-\omega_0)\leq\cos\!\left(\frac{\pi}{K}\right)\hat r_0 .
\label{eq:grid_inequality}
\end{equation}
For the true covariance these entries are $r_0=b+\sigma^2$ and $r_1=be^{i\omega_0}$. If $\hat{\mtx{C}}=\mtx{C}$, then \eqref{eq:grid_inequality} gives $b\leq\cos(\pi/K)(b+\sigma^2)$. Hence, if $b/(b+\sigma^2)>\cos(\pi/K)$, no choice of non-negative amplitudes on the grid can represent $\mtx{C}$, even with $K>P$ atoms and in the population setting. The covariances that the grid can represent are the non-negative combinations of $K$ fixed atoms, so they form a closed set, and the error of amplitude-only optimization then stays strictly positive.

The example depends on the grid, and for a fixed covariance the error can be made small by refining the grid. Still, every uniform grid fails on some covariance, and this motivates optimizing the frequencies together with the amplitudes.

\section{Benign Landscape of Joint Amplitude-Frequency Optimization}
\label{sec:jointTheory}

We work in the population setting, where the sample covariance is replaced by the true covariance. Since $\mtx{S}\to\mtx{C}$ almost surely as the number of samples grows, this is the limit of the finite-sample objective, and it separates the optimization problem from the noise in the data. Setting $\mtx{S}=\mtx{C}$ in \eqref{eq:objective} gives the population NLL
\begin{equation}
\begin{aligned}
L(\vct{a},\vct{\omega})
=
\operatorname{tr}\!\left(
\mtx{C}\hat{\mtx{C}}(\vct{a},\vct{\omega})^{-1}
\right)
+
\log\det \hat{\mtx{C}}(\vct{a},\vct{\omega}),
\end{aligned}
\label{eq:joint_nll_population}
\end{equation}
where $\mtx{C}$ is the true positive definite Toeplitz covariance matrix and $\hat{\mtx{C}}(\vct{a},\vct{\omega})$ is the model \eqref{eq:param_model} with $\omega_k\in\mathbb{R}$. The objective is defined on the open feasible set
\begin{equation}
    {\mathcal{F}}=\left\{(\vct{a},\vct{\omega})\;:\;\hat{\mtx{C}}(\vct{a},\vct{\omega})\succ0,\quad a_k>0\;k=1,\cdots,K\right\} .
\end{equation}
A \emph{parameter stationary point} of $L$ is a point $(\vct{a},\vct{\omega})\in {\mathcal{F}}$ for which all partial derivatives vanish, that is, $\nabla_{\vct{a}}L=\vct{0}$ and $\nabla_{\vct{\omega}}L=\vct{0}$. We refer to the set of stationary points of $L$ in $\mathcal{F}$ as the \emph{population landscape}, and we call it \emph{benign} if every stationary point satisfies $\hat{\mtx{C}}(\vct{a},\vct{\omega})=\mtx{C}$. Since $\operatorname{tr}(\mtx{C}\hat{\mtx{C}}^{-1})+\log\det\hat{\mtx{C}}$ is uniquely minimized over positive definite matrices at $\hat{\mtx{C}}=\mtx{C}$, a benign landscape means that every stationary point is a global minimizer, with no spurious local minima or saddle points to get stuck in.

\begin{theorem}
\label{theoremJointBenign}
Assume that $\mtx{C}\in\mathbb{C}^{P\times P}$ is Hermitian positive definite and Toeplitz.
Let $(\hat{\vct{a}},\hat{\vct{\omega}})\in {\mathcal{F}}$ be a stationary point of $L(\vct{a},\vct{\omega})$. Then $\hat{\mtx{C}}(\hat{\vct{a}},\hat{\vct{\omega}})=\mtx{C}$.
\end{theorem}

\begin{proof}
The proof uses two auxiliary lemmas, stated and proved in Appendix~\ref{app:lemmas}. The idea is that positive definiteness forces at least $P$ distinct frequencies, while stationarity makes each of them a double zero of a trigonometric polynomial of degree at most $P-1$, which is possible only if this polynomial vanishes identically.

Fix a stationary point $(\hat{\vct{a}},\hat{\vct{\omega}})$, write $\hat{\mtx{C}}=\hat{\mtx{C}}(\hat{\vct{a}},\hat{\vct{\omega}})$ and $\mtx{\Gamma}=\hat{\mtx{C}}-\mtx{C}$, and define the weighted residual
\begin{equation}
\mtx{E}:=\hat{\mtx{C}}^{-1}\mtx{\Gamma}\hat{\mtx{C}}^{-1}=\hat{\mtx{C}}^{-1}(\hat{\mtx{C}}-\mtx{C})\hat{\mtx{C}}^{-1},
\label{eq:joint_E_def}
\end{equation}
which is Hermitian. We show that $\mtx{\Gamma}=\mtx{0}$.

The derivatives of $\operatorname{tr}(\mtx{C}\hat{\mtx{C}}^{-1})$ and $\log\det\hat{\mtx{C}}$ in a Hermitian direction $\mtx{\Delta}$ are $-\operatorname{tr}(\hat{\mtx{C}}^{-1}\mtx{C}\hat{\mtx{C}}^{-1}\mtx{\Delta})$ and $\operatorname{tr}(\hat{\mtx{C}}^{-1}\mtx{\Delta})$, so the directional derivative of the population NLL with respect to $\hat{\mtx{C}}$ is $\operatorname{tr}(\mtx{E}\mtx{\Delta})$. The matrix $\hat{\mtx{C}}$ depends on $a_k$ and $\omega_k$ only through the term $a_k\vct{v}(\omega_k)\vct{v}(\omega_k)^H$. Hence, with $\mtx{E}$ held fixed and
\begin{equation}
p(\omega):=\vct{v}(\omega)^H\mtx{E}\vct{v}(\omega),
\label{eq:joint_dual_poly_def}
\end{equation}
the chain rule gives the following parameter derivatives of $L$ at the stationary point
\begin{align}
\frac{\partial L}{\partial a_k}&=p(\hat\omega_k),
\label{eq:joint_grad_a}\\
\frac{\partial L}{\partial \omega_k}&=\hat a_k\,p'(\hat\omega_k),
\label{eq:joint_grad_w}
\end{align}
where, since $\vct{v}'(\omega)=i\mtx{D}\vct{v}(\omega)$ with $\mtx{D}:=\operatorname{diag}(0,1,\ldots,{P-1})$,
\begin{equation}
p'(\omega)=-2\operatorname{Im}\bigl(\vct{v}(\omega)^H\mtx{E}\mtx{D}\vct{v}(\omega)\bigr).
\label{eq:joint_p_derivative}
\end{equation}
Since the point is stationary and $\hat a_k>0$,
\begin{equation}
p(\hat\omega_k)=0,\qquad p'(\hat\omega_k)=0,\qquad k=1,\ldots,K.
\label{eq:joint_double_zero_conditions}
\end{equation}

Expanding \eqref{eq:joint_dual_poly_def} entrywise, with indices $m,n=0,\ldots,P-1$, gives
\begin{equation}
\begin{aligned}
p(\omega)
&=\sum_{m,n=0}^{P-1}E_{m,n}e^{i(n-m)\omega}
=\sum_{\ell=-(P-1)}^{P-1}c_\ell e^{i\ell\omega},
\end{aligned}
\label{eq:joint_poly_expansion}
\end{equation}
where $c_\ell=\sum_m E_{m,m+\ell}$ is the sum of the entries of $\mtx{E}$ on its $\ell$-th diagonal, over the indices with $0\leq m,m+\ell\leq P-1$. Since $\mtx{E}$ is Hermitian, $c_{-\ell}=\overline{c_\ell}$, so $p$ is a real trigonometric polynomial of degree at most $P-1$.

Since $\vct{v}$, $p$ and $p'$ are $2\pi$-periodic, we may replace each $\hat\omega_k$ by its representative in $[0,2\pi)$ without changing $\hat{\mtx{C}}$ or the conditions \eqref{eq:joint_double_zero_conditions}, and we keep the same notation. Let $\widetilde\omega_1,\ldots,\widetilde\omega_r$ be the distinct values among $\hat\omega_1,\ldots,\hat\omega_K$. Merging the amplitudes of atoms with equal frequencies writes $\hat{\mtx{C}}$ as a sum of $r$ rank-one matrices, and since $\hat{\mtx{C}}\succ0$,
\begin{equation}
P=\operatorname{rank}(\hat{\mtx{C}})\leq r.
\label{eq:joint_r_geq_P}
\end{equation}
By \eqref{eq:joint_double_zero_conditions}, every $\widetilde\omega_i$ is a double zero of $p$. If $p$ were not identically zero, Lemma~\ref{lemmaTrigZeros} would give $r\leq P-1$, which contradicts \eqref{eq:joint_r_geq_P}. Hence $p\equiv0$, and since the coefficients of a trigonometric polynomial are unique,
\begin{equation}
c_\ell=0,\qquad \ell=-(P-1),\ldots,P-1.
\label{eq:joint_all_cl_zero}
\end{equation}
Every diagonal sum of $\mtx{E}$ is therefore zero, and Lemma~\ref{lemmaToeplitzOrthogonality} shows that $\mtx{E}$ is orthogonal to every Hermitian Toeplitz matrix.

Finally, $\mtx{\Gamma}=\hat{\mtx{C}}-\mtx{C}$ is Hermitian Toeplitz, so $\langle\mtx{\Gamma},\mtx{E}\rangle_F=0$. By \eqref{eq:joint_E_def} and the cyclic property of the trace,
\begin{equation}
\begin{aligned}
0=\langle\mtx{\Gamma},\mtx{E}\rangle_F
&=\operatorname{tr}\bigl(\mtx{\Gamma}\hat{\mtx{C}}^{-1}\mtx{\Gamma}\hat{\mtx{C}}^{-1}\bigr)\\
&=\bigl\|\hat{\mtx{C}}^{-1/2}\mtx{\Gamma}\hat{\mtx{C}}^{-1/2}\bigr\|_F^2 .
\end{aligned}
\end{equation}
Since $\hat{\mtx{C}}^{-1/2}$ is invertible, $\mtx{\Gamma}=\mtx{0}$, that is, $\hat{\mtx{C}}=\mtx{C}$.
\end{proof}

The theorem assumes $\hat{\mtx{C}}\succ0$ because $L$ is not defined otherwise. This is not restrictive for descent methods. Since $\mtx{C}\succ0$, Step~1 of Appendix~\ref{app:gdproof} shows that $L$ tends to infinity when the smallest eigenvalue of $\hat{\mtx{C}}$ tends to zero, so a method that starts from a positive definite $\hat{\mtx{C}}$ and never increases $L$, such as gradient descent with a backtracking line search, keeps $\hat{\mtx{C}}$ positive definite, with eigenvalues bounded away from zero. The theorem also holds for every $K\geq P$, including the minimal $K=P$, so it does not by itself favor overparameterization. The advantage of $K>P$ that we observe in Section~\ref{sec:experiments} lies in the speed of the optimization.

The proof shows that stationarity in our amplitude-frequency parameters implies stationarity in the covariance lags, both in the population setting and with finite samples. For Toeplitz covariances, stationarity in the linear lag coordinates, as in \cite{zwiernik2017maximum}, means that $\mtx{E}$ is orthogonal to every Hermitian Toeplitz direction. In our parameters $(\vct{a},\vct{\omega})$, stationarity only gives orthogonality to the directions generated by varying the amplitudes and frequencies. These directions depend on the point and need not span all Hermitian Toeplitz directions, and with fewer than $P$ distinct frequencies they cannot. The proof shows that they do span all of them whenever the amplitudes are positive and $\hat{\mtx{C}}\succ0$.

For finite samples, the same argument applied to a stationary point of \eqref{eq:objective} in $\mathcal{F}$, with $\mtx{E}=\hat{\mtx{C}}^{-1}(\hat{\mtx{C}}-\mtx{S})\hat{\mtx{C}}^{-1}$, shows that $\hat{\mtx{C}}$ is a stationary point of the NLL \eqref{eq:nll_C} over $\mathcal{T}$. Since $\hat{\mtx{C}}-\mtx{S}$ need not be Toeplitz, the last step of the population proof does not apply, and $\hat{\mtx{C}}$ need not be a global minimizer. Theorem~\ref{theoremNearPopulation} shows that when $\mtx{S}$ is close enough to $\mtx{C}$, the NLL has a unique stationary point over $\mathcal{T}$, which is the global MLE.

\section{Maximum Likelihood via Gradient Descent} \label{sec:gd}

We now turn the analysis of Section~\ref{sec:jointTheory} into a practical algorithm. To minimize the objective \eqref{eq:objective} by gradient descent, we enforce the constraint $a_k\geq0$ through a softplus reparameterization:
\begin{equation}
\hat{\mtx{C}}(\hat{\vct{a}},\hat{\vct{\omega}}) = \sum_{k=1}^{K} s(\hat{a}_k) \vct{v}(\hat{\omega}_k) \vct{v}(\hat{\omega}_k)^{\mathrm{H}},
\label{eq:caratheodory}
\end{equation}
where $s(\hat{a}_k) = \log(1 + e^{\hat{a}_k})$ is the softplus function, which maps the unconstrained parameters $\hat{\vct{a}}\in\mathbb{R}^K$ to positive amplitudes. With $\vct{\theta}=(\hat{\vct{a}},\hat{\vct{\omega}})$, we minimize $\mathcal{L}(\vct{\theta})=\mathrm{NLL}(\hat{\mtx{C}}(\vct{\theta}))$ by gradient descent, using the initialization in Algorithm~\ref{alg2}. Gradients of $\mathcal{L}$ are taken with respect to these unconstrained parameters.

Algorithm~\ref{alg2} starts from the positive definite matrix $\hat{\mtx{C}}^{(0)}=\frac{\operatorname{tr}(\mtx{S})}{P}\mtx{I}$, and its line search rejects trial steps at which $\hat{\mtx{C}}$ is singular by setting $\mathcal{L}=+\infty$ there. Every covariance iterate is therefore positive definite. Since $s'(\hat a_k)>0$, every stationary point of $\mathcal{L}$ is also a stationary point of $L$, so the finite-sample extension of Theorem~\ref{theoremJointBenign} applies.

The gradients follow from the proof of Theorem~\ref{theoremJointBenign}, with the softplus chain rule and $\mtx{S}$ in place of $\mtx{C}$:
\begin{align}
\frac {\partial \mathcal{L} }{\partial {\hat{a}_k}}&=s'(\hat{a}_k)\vct{v}_k^{\mathrm{H}} \mtx{E} \vct{v}_k,\\
\frac {\partial \mathcal{L} }{\partial {\hat{\omega}_k}} &= -2s(\hat{a}_k)\, {\rm{Im}}\left\{ \vct{v}_k^{\mathrm{H}} \mtx{E} \mtx{D} \vct{v}_k \right\}
\end{align}
where $\vct{v}_k=\vct{v}(\hat\omega_k)$, $\mtx{E}=\hat{\mtx{C}}^{-1}(\hat{\mtx{C}}-\mtx{S})\hat{\mtx{C}}^{-1}$ with $\hat{\mtx{C}}=\hat{\mtx{C}}(\hat{\vct{a}},\hat{\vct{\omega}})$, $\mtx{D}=\operatorname{diag}(0,1,\ldots,P-1)$, and $s'$ is the sigmoid function.

In early experiments, GD with separate amplitude and frequency step sizes converged much faster than GD with the single step size used in our conference version \cite{busbib2025toeplitz}. The two blocks of parameters have different sensitivities: the amplitudes $s(\hat{a}_k)$ enter $\hat{\mtx{C}}$ linearly, while $\vct{v}'(\omega)=i\mtx{D}\vct{v}(\omega)$ has entries as large as $P-1$. Using distinct step sizes for different blocks of variables is a standard approach \cite{peng2024block,beck2013block}. Algorithm~\ref{alg2} therefore scales the gradient by the fixed matrix $\mtx{\Lambda}=\operatorname{diag}(\eta^{(a)}_0\mtx{I}_K,\eta^{(\omega)}_0\mtx{I}_K)$ and uses an Armijo backtracking line search. With $\vct{\theta}^{(t)}=(\hat{\vct{a}}^{(t)},\hat{\vct{\omega}}^{(t)})$ and $\vct{g}^{(t)}=\nabla_{\vct{\theta}}\mathcal{L}(\vct{\theta}^{(t)})$, each iteration tries the steps $q=1,\beta,\beta^2,\ldots$ and accepts the first one that satisfies
\begin{equation}
\mathcal{L}\bigl(\vct{\theta}^{(t)}-q\mtx{\Lambda}\vct{g}^{(t)}\bigr)\leq\mathcal{L}\bigl(\vct{\theta}^{(t)}\bigr)-\alpha q\,\vct{g}^{(t)\top}\mtx{\Lambda}\vct{g}^{(t)},
\label{eq:armijo-split}
\end{equation}
where $\alpha\in(0,0.5)$ and $\beta\in(0,1)$ are fixed. Both step sizes are thus reduced by the same factor, and their ratio stays fixed.

Let
\begin{equation}
d=\bigl\|\mtx{C}^{-1/2}(\mtx{S}-\mtx{C})\mtx{C}^{-1/2}\bigr\|_F
\label{eq:near_pop_d}
\end{equation}
be the normalized distance between the sample covariance and the true covariance $\mtx{C}$.

\begin{theorem}
\label{theoremNearPopulation}
If $d<1/4$, then the NLL \eqref{eq:nll_C} has a unique stationary point over $\mathcal{T}$, which is its global minimizer $\hat{\mtx{C}}_{\mathrm{ML}}$. Moreover, for every $K\geq P$, the covariances $\hat{\mtx{C}}^{(t)}$ produced by Algorithm~\ref{alg2}, run without a limit on the number of iterations or backtracking steps, converge to $\hat{\mtx{C}}_{\mathrm{ML}}$.
\end{theorem}

In particular, the parameterized objective then has no spurious local minima: every stationary point of $\mathcal{L}$ whose covariance is positive definite gives $\hat{\mtx{C}}_{\mathrm{ML}}$. In the population setting, $\mtx{S}=\mtx{C}$, so $d=0$ and $\hat{\mtx{C}}_{\mathrm{ML}}=\mtx{C}$. The covariances $\hat{\mtx{C}}^{(t)}$ therefore converge to the true covariance.

To prove Theorem~\ref{theoremNearPopulation}, Appendix~\ref{app:nearpop} first shows that $d<1/4$ implies $\mtx{S}\succ0$ and places every stationary point of the NLL in $\{\mtx{X}\in\mathcal{T}:\mtx{X}\prec2\mtx{S}\}$, where the NLL is strictly convex in the covariance variable \cite{zwiernik2017maximum,boyd2004convex}. Together with existence of a minimizer for $\mtx{S}\succ0$, this establishes a unique stationary point, which is the global MLE. Lemma~\ref{lemmaGDLimit} in Appendix~\ref{app:gdproof} then shows that the covariance iterates generated by gradient descent stay in a compact set and that every limit point is stationary for the NLL over $\mathcal{T}$. Every limit point therefore equals this MLE, so the covariance sequence converges to it.

Since $\mtx{S}\to\mtx{C}$ as the number of samples grows, $d$ tends to zero and the theorem applies for large enough $M$. This requirement is conservative, and in all our finite-sample experiments $d>1$.

\begin{algorithm}[ht]
\caption{GD for Toeplitz Covariance Matrix Estimation}
\label{alg2}
\begin{algorithmic}[1]
\State \textbf{Input:} Sample covariance $\mtx{S}$ with $\operatorname{tr}(\mtx{S})>0$, parameter $K\geq P$, initial step sizes $\eta^{(a)}_0,\eta^{(\omega)}_0>0$, backtracking parameters $\alpha\in(0,0.5)$, $\beta\in(0,1)$, number of iterations $T$
\State \textbf{Initialize:} 

$\hat{\vct{\omega}}^{(0)} = 2\pi \cdot \left[0, \frac 1K, \frac 2K, \cdots, \frac{K-1}{K} \right]^\top$

$s(\hat{a}^{(0)}_k) = \frac{\operatorname{tr}(\mtx{S})}{PK}$, $k=1,\ldots,K$

\For{$t = 0, 1, 2, \dots, T-1$}
    \State Compute gradients:
    \[
        \nabla_{\hat{\vct{a}}} \mathcal{L}(\hat{\vct{a}}^{(t)}, \hat{\vct{\omega}}^{(t)}), \quad \nabla_{\hat{\vct{\omega}}} \mathcal{L}(\hat{\vct{a}}^{(t)}, \hat{\vct{\omega}}^{(t)})
    \]
    \State Initialize step sizes: $\eta^{(a)}_t = \eta^{(a)}_0$, $\eta^{(\omega)}_t = \eta^{(\omega)}_0$
    \Repeat
        \State Tentative update:
        \[
            \hat{\vct{a}}^{\mathrm{temp}} = \hat{\vct{a}}^{(t)} - \eta^{(a)}_t \nabla_{\hat{\vct{a}}} \mathcal{L}(\hat{\vct{a}}^{(t)}, \hat{\vct{\omega}}^{(t)}),
        \]
        \[
            \hat{\vct{\omega}}^{\mathrm{temp}} = \hat{\vct{\omega}}^{(t)} - \eta^{(\omega)}_t \nabla_{\hat{\vct{\omega}}} \mathcal{L}(\hat{\vct{a}}^{(t)}, \hat{\vct{\omega}}^{(t)})
        \]
        \State Check Armijo condition (\ref{eq:armijo-split})
        \If{Not satisfied}
            \State Backtrack: $\eta^{(a)}_t \leftarrow \beta \cdot \eta^{(a)}_t$, $\eta^{(\omega)}_t \leftarrow \beta \cdot \eta^{(\omega)}_t$
        \EndIf
    \Until{Armijo condition is satisfied}
    \State Accept update:
    \[
        \hat{\vct{a}}^{(t+1)} = \hat{\vct{a}}^{\mathrm{temp}}, \quad \hat{\vct{\omega}}^{(t+1)} = \hat{\vct{\omega}}^{\mathrm{temp}}
    \]
\EndFor
\State \textbf{Output:} Covariance estimate $\hat{\mtx{C}}(\hat{\vct{a}}^{(T)},\hat{\vct{\omega}}^{(T)})$
\end{algorithmic}
\end{algorithm}

The computational cost of the algorithm has two parts: the cost of one iteration and the number of iterations. Each iteration computes the matrix $\mtx{E}$, and each trial step of the line search computes $\mathcal{L}$, that is, $\operatorname{tr}(\mtx{S}\hat{\mtx{C}}^{-1})$ and $\log|\hat{\mtx{C}}|$. Our implementation forms $\hat{\mtx{C}}$ from its $K$ atoms, computes its inverse, and multiplies this inverse by the sample covariance $\mtx{S}$, which need not be Toeplitz, so each trial step costs $\mathcal{O}(P^3+KP^2)$. Given $\mtx{E}$, the gradients are cheap. As in the proof of Theorem~\ref{theoremJointBenign}, the partial derivatives are $p(\hat\omega_k)$ and $p'(\hat\omega_k)$, multiplied by $s'(\hat a_k)$ and $s(\hat a_k)$, and $p$ is determined by the $2P-1$ diagonal sums of $\mtx{E}$ in \eqref{eq:joint_poly_expansion}. Computing these sums costs $\mathcal{O}(P^2)$, and evaluating $p$ and $p'$ at the $K$ frequencies costs $\mathcal{O}(KP)$. For large $P$, the trial steps can be made cheaper using the Toeplitz structure of $\hat{\mtx{C}}$. For example, the Levinson--Durbin recursion gives $\log|\hat{\mtx{C}}|$ and the Trench algorithm gives $\hat{\mtx{C}}^{-1}$ in $\mathcal{O}(P^2)$ operations \cite{golub2013matrix, kailath1999fast, stoica2005spectral}. Overall, for a fixed number of line-search trials, the cost of an iteration grows only linearly with $K$.

The main factor in the run time is the number of iterations, which depends on the initialization, the step sizes, and the smoothness of the objective. In some Toeplitz instances we observed slow convergence, and the experiments show that overparameterization reduces the number of iterations.

\section{Experimental Results}\label{sec:experiments}

In this section, we report numerical experiments\footnote{The code for all experiments is available at \url{https://github.com/wiesel-lab/toeplitz-caratheodory}}. The first experiment is a single illustrative example. The second experiment uses $100$ random covariances in the population setting, and the third uses $320$ random finite-sample data sets.

We note that the goal of this section is to understand the optimization landscape of the Carath\'eodory decomposition and not to propose the fastest Toeplitz covariance solver. For a detailed comparison of GD versus ATOM \cite{ATOM24} we refer readers to the conference version of this work \cite{busbib2025toeplitz}, where the GD method reached accuracy comparable to or better than ATOM. Second-order methods can be much faster than GD. In Table~\ref{tab:perK}, for example, the Fisher scoring reference of Section~\ref{sec:finite} is more than ten times faster than GD. Future work may consider combining the Carath\'eodory parameterization with second-order methods such as the Newton algorithm of \cite{cederberg2024toeplitz}.

We evaluate Algorithm~\ref{alg2} at several overparameterization levels $K=F\cdot P$, and an amplitude-only variant that optimizes only the amplitudes on a fixed frequency grid. In all experiments, GD uses the backtracking parameters $\alpha = 0.3$ and $\beta = 0.5$, and the frequencies are initialized on the uniform grid as in Algorithm~\ref{alg2}. The iteration cap and the stopping rule differ between the experiments and are stated with each one. These are numerical choices, while our convergence results concern Algorithm~\ref{alg2} as stated, with no early stopping and no limit on the number of backtracking steps. All computations use double precision in Python with NumPy and Apple's Accelerate library, and all random draws use fixed seeds. Run times are elapsed wall-clock times on an Apple M3 Pro laptop, with $10$ runs executed in parallel, and for each finite-sample data set the reference and gradient descent are timed in the same process.

In the finite-sample experiment, we measure performance in terms of the mean squared error (MSE) of the first row of the covariance matrix,
\begin{equation}
\mathrm{MSE} \;=\; \frac{1}{P}\sum_{i=1}^P \left| \widehat{C}_{1i} - C_{1i} \right|^2,
\end{equation}
and compare it to the Cram\'er--Rao bound (CRB) for Toeplitz matrices as defined in \cite{ATOM24, turmon1994maximum, stoica2005spectral, kay1993estimation}. Specifically, let $\phi_1,\ldots,\phi_{2P-1}$ be $C_{11}$ and the real and imaginary parts of $C_{12},\ldots,C_{1P}$, and let $\mtx{J}$ be the Fisher information matrix, with entries $J_{ij}=M\operatorname{Re}\operatorname{tr}(\mtx{C}^{-1}\frac{\partial\mtx{C}}{\partial\phi_i}\mtx{C}^{-1}\frac{\partial\mtx{C}}{\partial\phi_j})$. For unbiased estimators, the expected MSE is bounded below by $\operatorname{tr}(\mtx{J}^{-1})/P$, and we use this value as a benchmark. Performance can also be measured using the Kullback-Leibler (KL) divergence \cite{Busbib2025SSP}. Our experiments with both metrics led to similar conclusions, and for compatibility with previous works we only report the MSE.

\subsection{Amplitude-Only Failure and the Effect of Joint Optimization}

We first illustrate the grid example of Section~\ref{sec:grid} in the population limit ($\mtx{S}=\mtx{C}$). We set $b=3$ and choose $\sigma^2$ so that the amplitude ratio is $b/(b+\sigma^2)=(1+\cos(\pi/K))/2$, strictly above the threshold $\cos(\pi/K)$. Note that both $\omega_0$ and $\sigma^2$ depend on $K$, so each $K$ has its own true covariance. We run Algorithm~\ref{alg2} in two stages: Stage~1 optimizes only the amplitudes on the fixed grid $\{2\pi(k-1)/K\}_{k=1}^{K}$, and Stage~2 starts from the final Stage~1 iterate and jointly optimizes the amplitudes and frequencies. The initial step sizes are $\eta_0^{(a)}=0.03$ in Stage~1, and $\eta_0^{(a)}=1$ and $\eta_0^{(\omega)}=6\cdot10^{-3}$ in Stage~2. Stage~1 runs $2{,}500$ iterations and Stage~2 runs $2{,}000$ iterations, without early stopping. We use $P=9$ and $K\in\{P,2P,3P,4P\}$.

\begin{figure}[t]
    \centering
    \includegraphics[width=\linewidth]{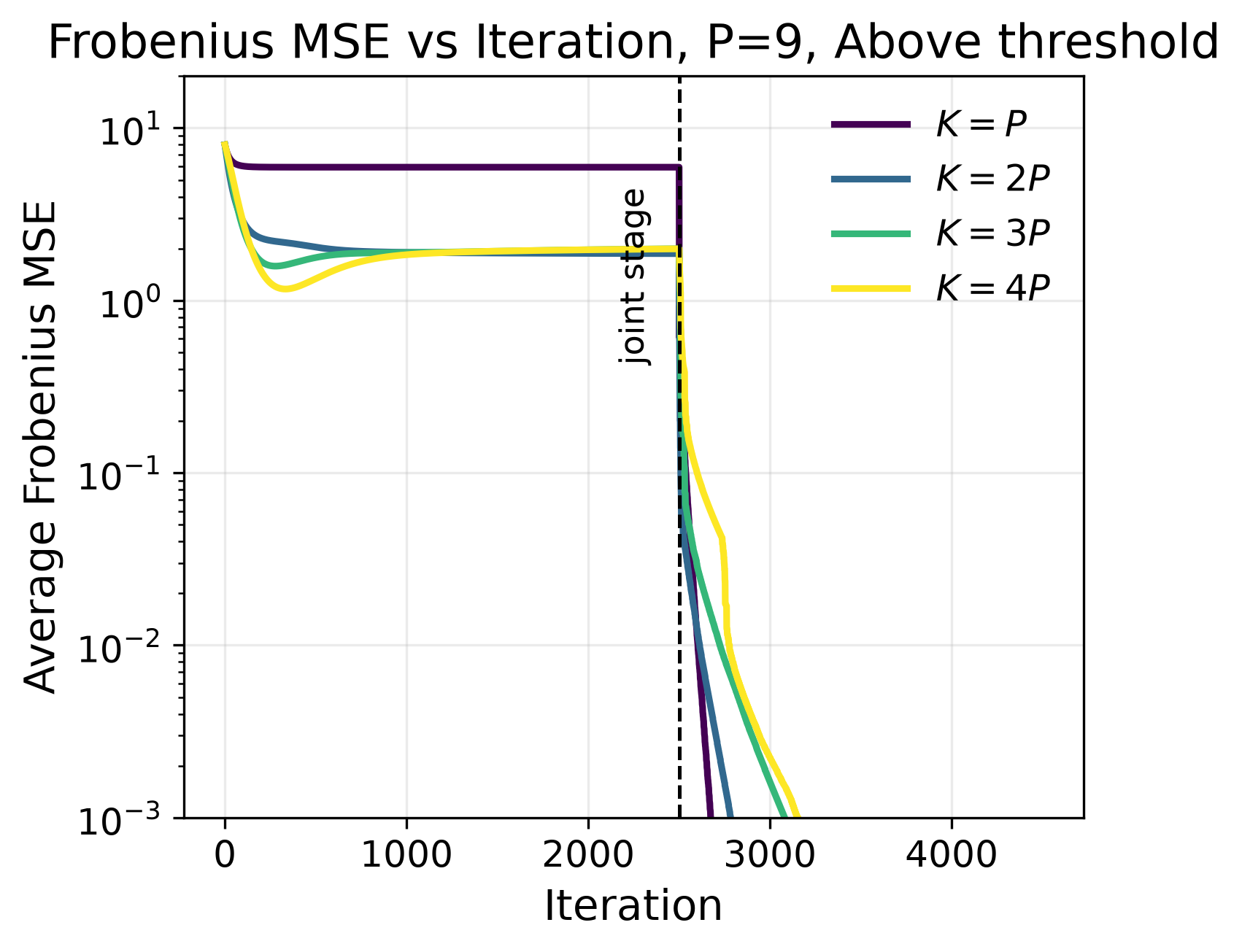}
    \caption{Mean squared entry error $\|\hat{\mtx{C}}-\mtx{C}\|_F^2/P^2$ versus iteration for the rank-one midpoint example ($P=9$), with the amplitude ratio strictly above the threshold $\cos(\pi/K)$. In Stage~1 (left of the dashed line), amplitude-only optimization on the fixed grid stays at a strictly positive error. In Stage~2, jointly optimizing the frequencies drives the error below $10^{-3}$ for every $K$, including $K=P$.}
    \label{fig:twostage}
\end{figure}

In Figure~\ref{fig:twostage}, the Stage~1 error cannot vanish even with more iterations, because \eqref{eq:grid_inequality} excludes an exact representation on the grid. Once Stage~2 begins, the error drops sharply, by at least three orders of magnitude, to below $10^{-3}$ within about $700$ iterations, for all $K$, including the minimal parameterization $K=P$. Running Stage~1 with the larger amplitude step size of Stage~2 gives nearly the same error levels, so this drop comes from moving the frequencies.

\subsection{Convergence versus Overparameterization}
\label{sec:success_kp}

\begin{figure*}[t]
    \centering
    \includegraphics[width=1.0\linewidth]{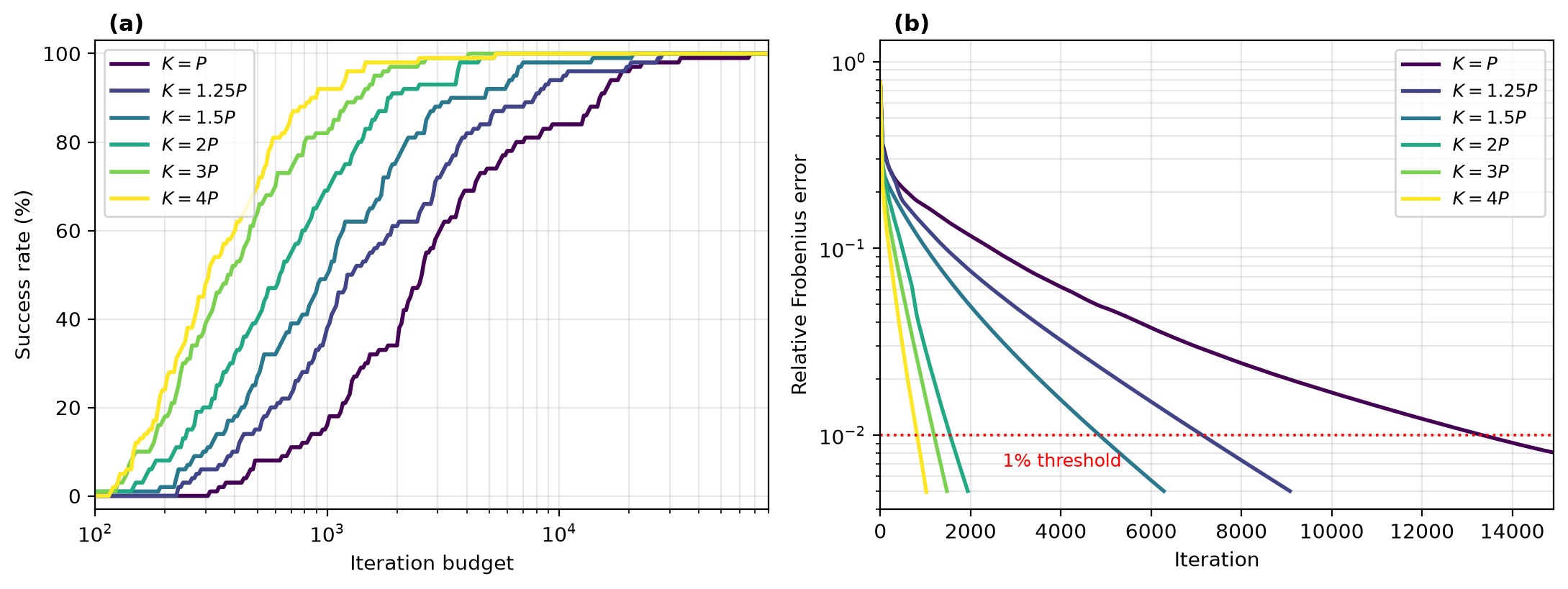}
    \caption{Convergence of joint gradient descent versus overparameterization in the population limit ($\mtx{S}=\mtx{C}$), over $50$ random full-rank covariances for each $P\in\{15,20\}$. (a) Success rate versus the iteration budget, that is, the fraction of runs reaching a relative Frobenius error below $10^{-2}$. All configurations eventually reach $100\%$, and overparameterization shifts the curves to the left. (b) Error trajectories for a single hard instance ($P=15$), with the same true and initial covariance for every $K$.}
    \label{fig:success_kp}
\end{figure*}

We study the effect of overparameterization on the convergence of joint gradient descent in the population limit, where $\mtx{S}=\mtx{C}$. Each true covariance is a full-rank Toeplitz matrix built from $P$ atoms plus the noise term $0.15\,\mtx{I}$. The frequencies of the atoms are drawn uniformly from $[0,2\pi)$, and their amplitudes uniformly from $[0.5,A]$, where $A$ is drawn uniformly from $[2,5]$ for each covariance. We use $P\in\{15,20\}$, overparameterization levels $K/P\in\{1,1.25,1.5,2,3,4\}$, and $50$ random covariances for each $P$. Each covariance is run with all six values of $K$, rounded to the nearest integer ($K=19$ and $22$ for $P=15$), for a total of $600$ runs. A run is declared successful once the relative Frobenius error $\|\hat{\mtx{C}}-\mtx{C}\|_F/\|\mtx{C}\|_F$ drops below $10^{-2}$. Each run is capped at $80{,}000$ iterations and stops once this error is below $5\cdot10^{-3}$. All runs are initialized as in Algorithm~\ref{alg2} and use the initial step sizes $\eta_0^{(a)}=0.3$ and $\eta_0^{(\omega)}=6\cdot10^{-3}$ for every $P$ and $K$.

In agreement with the theory, gradient descent succeeds in all $600$ runs, for every value of $K$. In Figure~\ref{fig:success_kp}(a), with a budget of $2{,}500$ iterations, $K=4P$ succeeds in $99\%$ of its $100$ runs, pooled over both values of $P$, while $K=P$ succeeds in only $48\%$. The median number of iterations drops from about $2{,}550$ at $K=P$ to about $310$ at $K=4P$, and the iteration count for the hardest instance drops from about $66{,}000$ to about $5{,}300$ iterations.

Figure~\ref{fig:success_kp}(b) illustrates this effect on a single hard instance, the second slowest of the $50$ instances with $P=15$ at $K=P$. Larger values of $K$ reach the threshold earlier, and $K=4P$ needs about one-sixteenth as many iterations as $K=P$. The cost of a trial step grows linearly with $K$, but the reduction in the number of iterations is much larger. The run time to the stopping error $5\cdot10^{-3}$ is shorter for $K=4P$ than for $K=P$ on all $100$ covariances. Across these covariances, the median ratio of the run time at $K=P$ to that at $K=4P$ is $9.5$.

\subsection{Finite-Sample Performance}
\label{sec:finite}

\begin{figure*}[t]
    \centering
    \includegraphics[width=\linewidth]{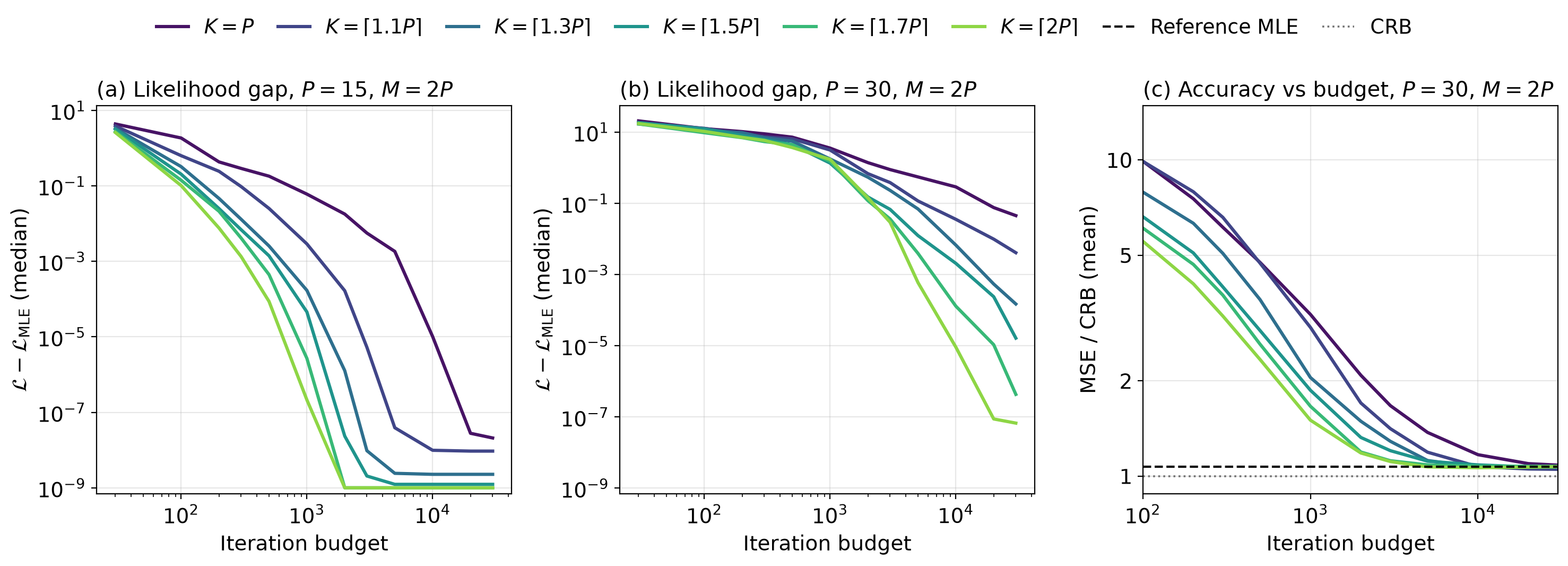}
    \caption{Finite-sample performance of joint gradient descent versus overparameterization, over $40$ random covariances for each $P$, with $M=2P$ and $K=\lceil FP\rceil$. (a,b) Median gap between the NLL reached by gradient descent and that of the reference MLE, versus the iteration budget, for $P=15$ and $P=30$. The gap decreases for every $K$, but larger $K$ closes it much faster, and the difference grows with the dimension. (c) Mean MSE normalized by the CRB versus the iteration budget, for $P=30$. The MSE for every $K$ decreases towards that of the reference MLE, which is close to the CRB, and larger $K$ gets there with fewer iterations.}
    \label{fig:expsuccess}
\end{figure*}

\begin{table*}[t]
\centering
\footnotesize
\caption{Per-$K$ diagnostics at the end of the runs, for $M=2P$ over $40$ data sets for each $P$. Iterations, time, $\mathcal{L}-\mathcal{L}_{\mathrm{MLE}}$ and $\|\nabla\mathcal{L}\|$ are medians, ``Capped'' is the fraction of runs that reached the cap of $30{,}000$ iterations, and MSE/CRB is the mean. Compared with $K=P$, $K=2P$ reduces the median number of iterations and the run time for both values of $P$, while the final MSE is similar for all $K$ and close to that of the reference MLE. The reference time includes both of its initializations.}
\label{tab:perK}
\begin{tabular}{lrrrrrr}
\toprule
 & Iterations & Capped & Time [s] & $\mathcal{L}-\mathcal{L}_{\mathrm{MLE}}$ & $\|\nabla\mathcal{L}\|$ & MSE/CRB\\
\midrule
\multicolumn{7}{l}{$P=15$, $M=30$}\\
$K=P$ & $15{,}415$ & $30\%$ & $5.54$ & $2.1\cdot10^{-8}$ & $2.9\cdot10^{-4}$ & $1.09$\\
$K=\lceil 1.1P\rceil$ & $5{,}673$ & $10\%$ & $2.17$ & $9.4\cdot10^{-9}$ & $1.9\cdot10^{-4}$ & $1.09$\\
$K=\lceil 1.3P\rceil$ & $3{,}150$ & $5\%$ & $1.01$ & $2.3\cdot10^{-9}$ & $1.7\cdot10^{-4}$ & $1.09$\\
$K=\lceil 1.5P\rceil$ & $2{,}310$ & $0\%$ & $0.74$ & $1.2\cdot10^{-9}$ & $1.1\cdot10^{-4}$ & $1.09$\\
$K=\lceil 1.7P\rceil$ & $1{,}754$ & $0\%$ & $0.56$ & $8.3\cdot10^{-10}$ & $8.8\cdot10^{-5}$ & $1.09$\\
$K=\lceil 2P\rceil$ & $1{,}425$ & $0\%$ & $0.44$ & $6.3\cdot10^{-10}$ & $8.9\cdot10^{-5}$ & $1.09$\\
Reference MLE & -- & -- & $0.03$ & $0$ & -- & $1.09$\\
\midrule
\multicolumn{7}{l}{$P=30$, $M=60$}\\
$K=P$ & $30{,}000$ & $98\%$ & $31.73$ & $4.5\cdot10^{-2}$ & $1.2\cdot10^{-1}$ & $1.08$\\
$K=\lceil 1.1P\rceil$ & $30{,}000$ & $90\%$ & $31.46$ & $4.1\cdot10^{-3}$ & $8.0\cdot10^{-2}$ & $1.05$\\
$K=\lceil 1.3P\rceil$ & $30{,}000$ & $65\%$ & $31.54$ & $1.5\cdot10^{-4}$ & $1.1\cdot10^{-2}$ & $1.07$\\
$K=\lceil 1.5P\rceil$ & $30{,}000$ & $60\%$ & $31.99$ & $1.6\cdot10^{-5}$ & $8.1\cdot10^{-3}$ & $1.07$\\
$K=\lceil 1.7P\rceil$ & $26{,}068$ & $50\%$ & $27.19$ & $4.3\cdot10^{-7}$ & $2.0\cdot10^{-3}$ & $1.07$\\
$K=\lceil 2P\rceil$ & $14{,}226$ & $38\%$ & $17.01$ & $6.7\cdot10^{-8}$ & $8.9\cdot10^{-4}$ & $1.07$\\
Reference MLE & -- & -- & $0.34$ & $0$ & -- & $1.07$\\
\bottomrule
\end{tabular}
\end{table*}

Each estimate is computed from the sample covariance $\mtx{S}$ of $M$ samples. The true covariances are drawn from the same random ensemble as in Section~\ref{sec:success_kp}, with $40$ covariances for each $P\in\{15,30\}$ and each $M\in\{P,2P,4P,8P\}$, for a total of $320$ data sets. We use the overparameterization factors $F\in\{1,1.1,1.3,1.5,1.7,2\}$ with $K=\lceil FP\rceil$. For every data set, all values of $K$ are run on the same $\mtx{S}$, with the same initial step sizes $\eta_0^{(a)}=0.3$ and $\eta_0^{(\omega)}=6\cdot10^{-3}$, and the initialization of Algorithm~\ref{alg2}. Each run is capped at $30{,}000$ iterations and stops earlier once the decrease of $\mathcal{L}$ is below $10^{-12}|\mathcal{L}|$ for $50$ consecutive iterations, or when no step satisfies the Armijo condition \eqref{eq:armijo-split} within $30$ trial steps, in which case the run returns its last accepted iterate.

We compute a numerical reference by Fisher scoring \cite{kay1993estimation} in the lag coordinates, started from two different initializations, a diagonally averaged sample covariance and a Burg estimate of order $P-1$, keeping the solution with the lower NLL.\footnote{The diagonally averaged matrix is made positive definite by shrinking it toward a multiple of the identity. The exact repair rule, line search and stopping tests of the reference solver are given with the code.} We call this solution the reference MLE and denote its NLL by $\mathcal{L}_{\mathrm{MLE}}$. Since the problem is nonconvex, this reference is not guaranteed to be the global optimum. In practice, the two initializations agreed to within $10^{-8}$ in NLL on $318$ of the $320$ data sets, and no gradient descent run reached an NLL below the reference by more than $5\cdot10^{-8}$.

Figures~\ref{fig:expsuccess}(a) and (b) show the NLL gap to the reference MLE, $\mathcal{L}-\mathcal{L}_{\mathrm{MLE}}$, as a function of the iteration budget for $M=2P$. The curves are ordered by $K$ at both dimensions. For $P=15$, the median NLL gap for $K=2P$ falls below $10^{-7}$ after about $2{,}000$ iterations, while $K=P$ requires about ten times as many iterations. Table~\ref{tab:perK} shows a similar reduction in the median run time. For $P=30$ the difference is larger: with $K=P$ almost all runs reach the cap of $30{,}000$ iterations, while with $K=2P$ the median gap is below $10^{-7}$. Pooling all sample sizes, the final gap is below $10^{-6}$ in $76\%$ and $100\%$ of the runs for $K=P$ and $K=2P$, respectively, at $P=15$. The corresponding fractions at $P=30$ are $2\%$ and $71\%$.

Since the CRB differs between covariances, we report the MSE normalized by the CRB of each covariance. The last column of Table~\ref{tab:perK} shows that at the end of the runs all values of $K$ give an MSE close to that of the reference MLE, even when the NLL gap has not fully closed. Figure~\ref{fig:expsuccess}(c) shows the MSE as a function of the budget for $P=30$ and $M=2P$: the normalized MSE for every $K$ decreases towards that of the reference MLE as the budget grows, and larger values of $K$ get there earlier.

\section{Discussion and Future Work}\label{sec:conclusions}

This paper studied the optimization landscape of Toeplitz covariance estimation under the Carath\'eodory decomposition. Fixed-grid amplitude-only optimization can fail under grid mismatch, even in the population setting, while joint amplitude-frequency optimization has a benign population landscape. With finite samples, every stationary point of the parameterized problem that produces a positive definite covariance maps to a stationary point of the Toeplitz maximum likelihood problem. The covariance iterates generated by gradient descent converge to the true covariance in the population setting, and to the unique maximum likelihood estimate when the sample covariance is close enough to the true covariance. Empirically, overparameterization mainly speeds up the optimization, and for a fixed number of iterations this yields more accurate estimates.

Future work will focus on characterizing the stationary points of the finite-sample problem when the sample covariance is not close to the true covariance, analyzing the rate of convergence of the covariance iterates and its dependence on $K$, and adapting the proposed method to block-Toeplitz structures for multi-dimensional array processing.

\appendices
\section{Auxiliary Lemmas}
\label{app:lemmas}

\begin{lemma}
\label{lemmaTrigZeros}
Let $p$ be a nonzero real trigonometric polynomial of degree at most $P-1$,
$p(\omega)=\sum_{\ell=-(P-1)}^{P-1}c_\ell e^{i\ell\omega}$.
If $\widetilde\omega_1,\ldots,\widetilde\omega_r\in[0,2\pi)$ are distinct and satisfy
$p(\widetilde\omega_i)=p'(\widetilde\omega_i)=0$ for $i=1,\ldots,r$,
then $r\leq P-1$.
\end{lemma}

\begin{proof}
Set $z=e^{i\omega}$, so that $z$ lies on the unit circle $|z|=1$ and $e^{i\ell\omega}=z^\ell$. Hence, on the unit circle,
\begin{equation}
\begin{aligned}
p(\omega)
=
\widetilde p(z),
\qquad
\widetilde p(z)
:=
\sum_{\ell=-(P-1)}^{P-1}
c_\ell z^\ell.
\end{aligned}
\end{equation}
The function $\widetilde p(z)$ is a Laurent polynomial, because it may contain negative powers of $z$.
Multiplying by $z^{P-1}$ removes all negative powers:
\begin{equation}
\begin{aligned}
q(z)
:=
z^{P-1}\widetilde p(z)
=
\sum_{\ell=-(P-1)}^{P-1}
c_\ell z^{\ell+P-1}.
\end{aligned}
\end{equation}
Thus $q(z)$ is an ordinary algebraic polynomial of degree at most $2(P-1)$. Since $p$ is nonzero, at least one coefficient $c_\ell$ is nonzero, and therefore $q$ is a nonzero polynomial.

Let $z_i=e^{i\widetilde\omega_i}$ for $i=1,\ldots,r$. Since the frequencies $\widetilde\omega_i$ are distinct points of $[0,2\pi)$, the points $z_i$ are distinct points on the unit circle, and in particular $z_i\neq0$. We show that each $z_i$ is a zero of $q$ of multiplicity at least two. First, $p(\widetilde\omega_i)=0$ gives $\widetilde p(z_i)=0$, and hence $q(z_i)=z_i^{P-1}\widetilde p(z_i)=0$. Second, differentiating $p(\omega)=\widetilde p(e^{i\omega})$ with the chain rule gives
\begin{equation}
p'(\omega)
=
ie^{i\omega}\,\widetilde p\,'(e^{i\omega}).
\end{equation}
Since $p'(\widetilde\omega_i)=0$ and $z_i\neq0$, we get $\widetilde p\,'(z_i)=0$. Therefore
\begin{equation}
q'(z_i)
=
(P-1)z_i^{P-2}\widetilde p(z_i)
+
z_i^{P-1}\widetilde p\,'(z_i)
=
0,
\end{equation}
so $(z-z_i)^2$ divides $q(z)$.

Since the points $z_i$ are distinct, the product $\prod_{i=1}^{r}(z-z_i)^2$ divides $q(z)$. Comparing the degrees of these nonzero polynomials gives $2r\leq 2(P-1)$, that is, $r\leq P-1$.
\end{proof}

\begin{lemma}
\label{lemmaToeplitzOrthogonality}
Let $\mtx{E}\in\C^{P\times P}$ be a Hermitian matrix and define its diagonal sums
\begin{equation}
c_\ell
:=
\sum_{\substack{m:\ 0\leq m\leq P-1 \\ 0\leq m+\ell\leq P-1}}
E_{m,m+\ell},
\qquad
\ell=-(P-1),\ldots,P-1.
\end{equation}
Assume that $c_\ell=0$ for all $\ell=-(P-1),\ldots,P-1$.
Then, for every Hermitian Toeplitz matrix $\mtx{T}$,
\begin{equation}
\begin{aligned}
\langle \mtx{T},\mtx{E}\rangle_F=0.
\end{aligned}
\end{equation}
\end{lemma}

\begin{proof}
Let $\mtx{T}$ be Hermitian Toeplitz, so there exist values $t_\ell$ with $T_{m,n}=t_{n-m}$.
The Frobenius inner product is the entrywise inner product of two matrices:
\begin{align}
\langle \mtx{T},\mtx{E}\rangle_F
&=
\operatorname{tr}(\mtx{T}^H\mtx{E})
\notag\\
&=
\sum_{m=0}^{P-1}
\sum_{n=0}^{P-1}
\overline{T_{m,n}}E_{m,n}
\notag\\
&=
\sum_{m=0}^{P-1}
\sum_{n=0}^{P-1}
\overline{t_{n-m}}E_{m,n}.
\label{eq:toeplitz-inner-product}
\end{align}
Grouping the terms according to the lag $\ell=n-m$, and using that all entries of $\mtx{T}$ on a fixed diagonal share the value $t_\ell$,
\begin{equation}
\begin{aligned}
\langle \mtx{T},\mtx{E}\rangle_F
=
\sum_{\ell=-(P-1)}^{P-1}
\overline{t_\ell}
\sum_{\substack{m:\ 0\leq m\leq P-1 \\ 0\leq m+\ell\leq P-1}}
E_{m,m+\ell}
=
\sum_{\ell=-(P-1)}^{P-1}
\overline{t_\ell}c_\ell.
\end{aligned}
\end{equation}
Since $c_\ell=0$ for every $\ell$, we obtain $\langle \mtx{T},\mtx{E}\rangle_F=0$.
\end{proof}

\section{Limit Points of Gradient Descent}
\label{app:gdproof}

\begin{lemma}
\label{lemmaGDLimit}
Assume that $\mtx{S}\succ0$, and run Algorithm~\ref{alg2} without a limit on the number of iterations or backtracking steps. Then the covariances $\hat{\mtx{C}}^{(t)}$ stay in a compact set of positive definite matrices, and every limit point of $\hat{\mtx{C}}^{(t)}$ is a stationary point of the NLL \eqref{eq:nll_C} over $\mathcal{T}$.
\end{lemma}

\begin{proof}
Write $\vct{\theta}=(\hat{\vct{a}},\hat{\vct{\omega}})$ for the parameters of Algorithm~\ref{alg2}, $\vct{\theta}^{(t)}$ for its iterates, and $\hat{\mtx{C}}^{(t)}=\hat{\mtx{C}}(\vct{\theta}^{(t)})$. We set $\mathcal{L}(\vct{\theta})=+\infty$ when $\hat{\mtx{C}}(\vct{\theta})$ is singular, and we write
\begin{equation}
c:=\lambda_{\min}(\mtx{S})>0,\qquad \mathcal{L}_0:=\mathcal{L}(\vct{\theta}^{(0)}).
\end{equation}
The proof has three steps.

\emph{Step 1: the eigenvalues stay in a fixed interval.}
Let $\hat\lambda_1,\ldots,\hat\lambda_P$ be the eigenvalues of a positive definite $\hat{\mtx{C}}$. Since $\mtx{S}\succeq c\mtx{I}$, we have $\operatorname{tr}(\mtx{S}\hat{\mtx{C}}^{-1})\geq c\operatorname{tr}(\hat{\mtx{C}}^{-1})$, and hence
\begin{equation}
\mathcal{L}
=\operatorname{tr}(\mtx{S}\hat{\mtx{C}}^{-1})+\log\det\hat{\mtx{C}}
\geq\sum_{i=1}^{P}f(\hat\lambda_i),
\quad
f(\lambda):=\frac{c}{\lambda}+\log\lambda .
\label{eq:gd_lower_bound}
\end{equation}
The function $f$ has minimum $1+\log c$ and tends to infinity as $\lambda\to0$ or $\lambda\to\infty$. Therefore, on the sublevel set $\mathcal{L}\leq\mathcal{L}_0$, each eigenvalue satisfies
\begin{equation}
f(\hat\lambda_i)\leq\mathcal{L}_0-(P-1)(1+\log c).
\end{equation}
Thus all eigenvalues lie in a fixed interval $[\lambda_{\mathrm{lo}},\lambda_{\mathrm{hi}}]$ with $0<\lambda_{\mathrm{lo}}\leq\lambda_{\mathrm{hi}}<\infty$. The line search keeps the covariance iterates in this sublevel set, so they lie in the compact set of Hermitian Toeplitz matrices satisfying $\lambda_{\mathrm{lo}}\mtx{I}\preceq\hat{\mtx{C}}\preceq\lambda_{\mathrm{hi}}\mtx{I}$. Finally, each atom has trace $P$, so $\operatorname{tr}(\hat{\mtx{C}})=P\sum_k s(\hat a_k)\leq P\lambda_{\mathrm{hi}}$. Every amplitude therefore satisfies $s(\hat a_k)\leq\lambda_{\mathrm{hi}}$.

\emph{Step 2: the parameter gradient tends to zero.}
Since both step sizes are reduced by the same factor, each iteration moves along the direction $-\mtx{\Lambda}\nabla_{\vct{\theta}}\mathcal{L}$, where $\mtx{\Lambda}=\operatorname{diag}(\eta^{(a)}_0\mtx{I}_K,\eta^{(\omega)}_0\mtx{I}_K)$ is fixed and positive definite. For this fixed positive definite scaling, the usual Armijo convergence argument requires the gradient of $\mathcal{L}$ to be Lipschitz continuous near the iterates \cite{bertsekas1999nonlinear}. We verify this condition. By Step~1, the eigenvalues of $\hat{\mtx{C}}^{(t)}$ are at least $\lambda_{\mathrm{lo}}$ and the amplitudes are at most $\lambda_{\mathrm{hi}}$. The map from the parameters to $\hat{\mtx{C}}$ is Lipschitz on such a set, since the softplus function is $1$-Lipschitz and $\vct{v}(\omega)$ is smooth and periodic. Hence a small change of the parameters cannot make $\hat{\mtx{C}}$ singular, and on a fixed neighborhood of the iterates the matrix $\hat{\mtx{C}}^{-1}$, the amplitudes, and all derivatives of the softplus function and of $\vct{v}(\omega)$ are bounded. The second derivatives of $\mathcal{L}$ are built from these quantities and from $\mtx{S}$, so they are bounded as well, and the condition holds. The gradient is also bounded on this neighborhood, so sufficiently small steps from an iterate stay inside it.

The line search never needs to reduce the step size below a fixed positive value, so each iteration decreases $\mathcal{L}$ by at least a fixed multiple of $\|\nabla_{\vct{\theta}}\mathcal{L}(\vct{\theta}^{(t)})\|^2$. Since $\mathcal{L}$ is bounded from below by \eqref{eq:gd_lower_bound}, the sum of $\|\nabla_{\vct{\theta}}\mathcal{L}(\vct{\theta}^{(t)})\|^2$ over all iterations is finite, and therefore
\begin{equation}
\nabla_{\vct{\theta}}\mathcal{L}(\vct{\theta}^{(t)})\to\vct{0}.
\label{eq:gd_grad_to_zero}
\end{equation}

\emph{Step 3: every covariance limit point is stationary for the NLL over $\mathcal{T}$.}
We now transfer vanishing parameter derivatives to stationarity along all Toeplitz directions. Let $\bar{\mtx{C}}$ be a limit point of $\hat{\mtx{C}}^{(t)}$. It lies in the compact set of Step~1, so $\bar{\mtx{C}}\succ0$. We would like to pass to the limit in the parameters as well, but they need not converge: an amplitude $s(\hat a_k)$ may tend to zero, in which case $\hat a_k\to-\infty$. We therefore take limits only of quantities that stay bounded. Since $\hat{\mtx{C}}$ depends on each frequency only through $e^{i\hat\omega_k}$, we may reduce the frequencies modulo $2\pi$ without changing anything. By Step~1 the amplitudes lie in $[0,\lambda_{\mathrm{hi}}]$. Hence there is a subsequence along which
\begin{equation}
\hat{\mtx{C}}^{(t)}\to\bar{\mtx{C}},\quad
\hat\omega_k\to\bar\omega_k,\quad
s(\hat a_k)\to\bar s_k\in[0,\lambda_{\mathrm{hi}}],
\end{equation}
for $k=1,\ldots,K$,
and then
\begin{equation}
\bar{\mtx{C}}=\sum_{k=1}^{K}\bar s_k\,\vct{v}(\bar\omega_k)\vct{v}(\bar\omega_k)^{\mathrm{H}}.
\label{eq:gd_limit_cov}
\end{equation}
The limits $\bar\omega_k$ lie in $[0,2\pi]$. Every quantity below depends on a frequency only through $\vct{v}$, which is $2\pi$-periodic, so a limit $\bar\omega_k=2\pi$ can be replaced by $0$, and we may take $\bar\omega_k\in[0,2\pi)$.
Since $\bar{\mtx{C}}$ is invertible, the residual matrices $\mtx{E}=\hat{\mtx{C}}^{-1}(\hat{\mtx{C}}-\mtx{S})\hat{\mtx{C}}^{-1}$ converge along the subsequence to
\begin{equation}
\bar{\mtx{E}}:=\bar{\mtx{C}}^{-1}(\bar{\mtx{C}}-\mtx{S})\bar{\mtx{C}}^{-1},
\end{equation}
and the polynomials $p(\omega)=\vct{v}(\omega)^{\mathrm{H}}\mtx{E}\vct{v}(\omega)$ of \eqref{eq:joint_dual_poly_def} and their derivatives converge uniformly in $\omega$ to $\bar p$ and $\bar p'$, respectively, where
\begin{equation}
\bar p(\omega):=\vct{v}(\omega)^{\mathrm{H}}\bar{\mtx{E}}\vct{v}(\omega).
\end{equation}

We now use \eqref{eq:gd_grad_to_zero}. As in the proof of Theorem~\ref{theoremJointBenign}, the partial derivatives of $\mathcal{L}$ are
\begin{equation}
\frac{\partial\mathcal{L}}{\partial\hat a_k}=s'(\hat a_k)\,p(\hat\omega_k),
\qquad
\frac{\partial\mathcal{L}}{\partial\hat\omega_k}=s(\hat a_k)\,p'(\hat\omega_k),
\end{equation}
and both tend to zero. Consider an atom whose amplitude does not vanish in the limit, $\bar s_k>0$. Its first factors do not vanish either: $s(\hat a_k)\to\bar s_k>0$, and since the softplus function satisfies $s'(\hat a)=1-e^{-s(\hat a)}$, also $s'(\hat a_k)\to1-e^{-\bar s_k}>0$. A product that tends to zero while its first factor stays away from zero must have its second factor tend to zero. Therefore
\begin{equation}
\bar p(\bar\omega_k)=0,\qquad \bar p'(\bar\omega_k)=0,\qquad\text{for every $k$ with $\bar s_k>0$},
\label{eq:gd_double_zero}
\end{equation}
which are exactly the conditions \eqref{eq:joint_double_zero_conditions} at a stationary point in the proof of Theorem~\ref{theoremJointBenign}. Atoms with $\bar s_k=0$ give no such condition, but they do not appear in \eqref{eq:gd_limit_cov} either. Since $\bar{\mtx{C}}\succ0$ is built only from the atoms with $\bar s_k>0$, these atoms have at least $P$ distinct frequencies, which is the condition \eqref{eq:joint_r_geq_P}. From this point on, the proof of Theorem~\ref{theoremJointBenign} applies word for word up to its last step: $\bar p\equiv0$ by Lemma~\ref{lemmaTrigZeros}, and then $\bar{\mtx{E}}$ is orthogonal to every Hermitian Toeplitz matrix by Lemma~\ref{lemmaToeplitzOrthogonality}. Since $\bar{\mtx{E}}$ is the gradient of the NLL \eqref{eq:nll_C} at $\bar{\mtx{C}}$, this means that $\bar{\mtx{C}}$ is a stationary point of \eqref{eq:nll_C} over $\mathcal{T}$, which is the second claim of the lemma.
\end{proof}

\section{Proof of Theorem~\ref{theoremNearPopulation}}
\label{app:nearpop}

Throughout the proof, $\mtx{C}$ is the true covariance, $\mtx{S}$ is the sample covariance, and we write
\begin{equation}
\mtx{W}:=\mtx{C}^{-1/2}(\mtx{S}-\mtx{C})\mtx{C}^{-1/2},
\qquad
d=\|\mtx{W}\|_F<\frac14 .
\end{equation}
Since $\|\mtx{W}\|_2\leq\|\mtx{W}\|_F=d$, every eigenvalue of $\mtx{I}+\mtx{W}$ is at least $1-d$, and therefore
\begin{equation}
\mtx{S}=\mtx{C}^{1/2}(\mtx{I}+\mtx{W})\mtx{C}^{1/2}\succeq(1-d)\,\mtx{C}\succ0 .
\label{eq:np_S_lower}
\end{equation}
Steps~1--3 analyze the NLL directly in the covariance variable over $\mathcal{T}$. Step~4 uses Lemma~\ref{lemmaGDLimit} to connect this analysis to the covariance iterates generated by GD in the Carath\'eodory parameters.

\emph{Step 1: every stationary point is close to $\mtx{C}$.}
Let $\hat{\mtx{C}}\in\mathcal{T}$ be a stationary point of \eqref{eq:nll_C} over $\mathcal{T}$. The gradient of the NLL at $\hat{\mtx{C}}$, $\mtx{E}=\hat{\mtx{C}}^{-1}(\hat{\mtx{C}}-\mtx{S})\hat{\mtx{C}}^{-1}$, is then orthogonal to every Hermitian Toeplitz matrix, and in particular to $\mtx{T}=\hat{\mtx{C}}-\mtx{C}$. We write $\hat{\mtx{C}}-\mtx{S}=(\hat{\mtx{C}}-\mtx{C})-(\mtx{S}-\mtx{C})$ and introduce the Hermitian matrices
\begin{align}
\mtx{X}&:=\hat{\mtx{C}}^{-1/2}(\hat{\mtx{C}}-\mtx{C})\hat{\mtx{C}}^{-1/2},\\
\mtx{Y}&:=\hat{\mtx{C}}^{-1/2}(\mtx{S}-\mtx{C})\hat{\mtx{C}}^{-1/2},
\end{align}
which measure the estimation error and the sampling error, both normalized by $\hat{\mtx{C}}$. Using the cyclic property of the trace, the orthogonality condition $\operatorname{tr}(\mtx{T}\mtx{E})=0$ becomes
\begin{equation}
\|\mtx{X}\|_F^2=\operatorname{tr}(\mtx{X}\mtx{Y})\leq\|\mtx{X}\|_F\,\|\mtx{Y}\|_F,
\end{equation}
by the Cauchy--Schwarz inequality, so $\|\mtx{X}\|_F\leq\|\mtx{Y}\|_F$. It remains to relate $\mtx{Y}$, which is normalized by $\hat{\mtx{C}}$, to $d$, which is normalized by $\mtx{C}$. With $\mtx{G}:=\hat{\mtx{C}}^{-1/2}\mtx{C}^{1/2}$ we have $\mtx{Y}=\mtx{G}\mtx{W}\mtx{G}^{\mathrm{H}}$, and therefore
\begin{equation}
\|\mtx{Y}\|_F\leq\|\mtx{G}\|_2^2\,\|\mtx{W}\|_F=m\,d,
\end{equation}
where $m:=\lambda_{\max}(\hat{\mtx{C}}^{-1/2}\mtx{C}\hat{\mtx{C}}^{-1/2})$.
On the other hand, $\mtx{X}=\mtx{I}-\hat{\mtx{C}}^{-1/2}\mtx{C}\hat{\mtx{C}}^{-1/2}$ has the eigenvalue $1-m$, so $\|\mtx{X}\|_F\geq m-1$. Putting the three inequalities together gives $m-1\leq\|\mtx{X}\|_F\leq m\,d$, hence $m\leq1/(1-d)$, and finally
\begin{equation}
\|\mtx{X}\|_F\leq\frac{d}{1-d}.
\label{eq:near_pop_X}
\end{equation}

\emph{Step 2: every stationary point lies in a convex region.}
The eigenvalues of $\hat{\mtx{C}}^{-1/2}\mtx{C}\hat{\mtx{C}}^{-1/2}=\mtx{I}-\mtx{X}$ differ from $1$ by at most $\|\mtx{X}\|_2\leq\|\mtx{X}\|_F$, so by \eqref{eq:near_pop_X} they are at least $1-d/(1-d)=(1-2d)/(1-d)$. Multiplying by $\hat{\mtx{C}}^{1/2}$ on both sides,
\begin{equation}
\hat{\mtx{C}}\preceq\frac{1-d}{1-2d}\,\mtx{C}.
\label{eq:np_C_upper}
\end{equation}
Together with \eqref{eq:np_S_lower}, this gives
\begin{align}
2\mtx{S}-\hat{\mtx{C}}&\succeq\left(2(1-d)-\frac{1-d}{1-2d}\right)\mtx{C}\nonumber\\
&=\frac{(1-d)(1-4d)}{1-2d}\,\mtx{C}\succ0,
\end{align}
where the last inequality is exactly where $d<1/4$ is used. Hence every stationary point of \eqref{eq:nll_C} over $\mathcal{T}$ lies in the set
\begin{equation}
\mathcal{D}:=\{\mtx{C}'\in\mathcal{T}:\ \mtx{C}'\prec2\mtx{S}\},
\end{equation}
which is convex, since it is the intersection of the Toeplitz subspace with two convex sets of matrices.

\emph{Step 3: existence and uniqueness.}
The NLL is strictly convex in the covariance variable on $\mathcal{D}$ \cite{zwiernik2017maximum}, \cite[Exercise~7.4(b)]{boyd2004convex}. The same argument holds for complex Hermitian matrices: for $\mtx{C}'\in\mathcal{D}$ and a nonzero Hermitian Toeplitz direction $\mtx{\Delta}$,
\begin{equation}
\begin{aligned}
&\left.\frac{d^2}{d\tau^2}\mathrm{NLL}(\mtx{C}'+\tau\mtx{\Delta})\right|_{\tau=0}\\
&\qquad=\operatorname{tr}\bigl(\mtx{A}(2\mtx{B}-\mtx{I})\mtx{A}\bigr)>0,
\end{aligned}
\end{equation}
where $\mtx{A}=\mtx{C}'^{-1/2}\mtx{\Delta}\mtx{C}'^{-1/2}$ and $\mtx{B}=\mtx{C}'^{-1/2}\mtx{S}\mtx{C}'^{-1/2}$. The inequality follows from $\mtx{A}\neq\mtx{0}$ and $2\mtx{B}-\mtx{I}=\mtx{C}'^{-1/2}(2\mtx{S}-\mtx{C}')\mtx{C}'^{-1/2}\succ0$.

Since $\mathcal{D}$ is convex and contains every stationary point, strict convexity gives at most one such point. Also, $\mtx{S}\succ0$, so the bound \eqref{eq:gd_lower_bound} in Appendix~\ref{app:gdproof} gives compact sublevel sets of the NLL over $\mathcal{T}$. The NLL therefore attains its global minimum at some $\hat{\mtx{C}}_{\mathrm{ML}}\in\mathcal{T}$. This minimizer is stationary along every Toeplitz direction, so it is the unique stationary point. This proves the first claim of the theorem.

\emph{Step 4: the covariance iterates converge to the MLE.}
By \eqref{eq:np_S_lower}, $\mtx{S}\succ0$, so Lemma~\ref{lemmaGDLimit} applies: every limit point of $\hat{\mtx{C}}^{(t)}$ is a stationary point of \eqref{eq:nll_C} over $\mathcal{T}$, and by Step~3 it equals $\hat{\mtx{C}}_{\mathrm{ML}}$. The covariances $\hat{\mtx{C}}^{(t)}$ stay in a compact set and have $\hat{\mtx{C}}_{\mathrm{ML}}$ as their only limit point, so they converge to $\hat{\mtx{C}}_{\mathrm{ML}}$.

\bibliographystyle{IEEEtran}
\bibliography{main}
\end{document}